\documentclass[nohyperref]{article}

\usepackage[subrefformat=parens]{subcaption}  

\usepackage{microtype}
\usepackage{graphicx}
\usepackage{booktabs} 

\usepackage{hyperref}

\usepackage[accepted]{icml2023}

\usepackage{amsmath}
\usepackage{amssymb}
\usepackage{mathtools}
\usepackage{amsthm}

\usepackage[capitalize,noabbrev]{cleveref}

\theoremstyle{plain}

\theoremstyle{definition}

\theoremstyle{remark}

\usepackage[textsize=tiny]{todonotes}

\icmltitlerunning{Cell-Free Latent Go-Explore}

\begin{document}

\twocolumn[
\icmltitle{Cell-Free Latent Go-Explore}



\icmlsetsymbol{equal}{*}

\begin{icmlauthorlist}
\icmlauthor{Quentin Gallouédec}{ecl}
\icmlauthor{Emmanuel Dellandréa}{ecl}
\end{icmlauthorlist}

\icmlaffiliation{ecl}{École Centrale de Lyon, LIRIS, CNRS UMR 5205, France}

\icmlcorrespondingauthor{Quentin Gallouédec}{quentin.gallouedec@ec-lyon.fr}

\icmlkeywords{Machine Learning, ICML}

\vskip 0.3in
]




\begin{abstract}
In this paper, we introduce Latent Go-Explore (LGE), a simple and general approach based on the Go-Explore paradigm for exploration in reinforcement learning (RL).
Go-Explore was initially introduced with a strong domain knowledge constraint for partitioning the state space into cells.
However, in most real-world scenarios, drawing domain knowledge from raw observations is complex and tedious.
If the cell partitioning is not informative enough, Go-Explore can completely fail to explore the environment.
We argue that the Go-Explore approach can be generalized to any environment without domain knowledge and without cells by exploiting a learned latent representation.
Thus, we show that LGE can be flexibly combined with any strategy for learning a latent representation.
Our results indicate that LGE, although simpler than Go-Explore, is more robust and outperforms state-of-the-art algorithms in terms of pure exploration on multiple hard-exploration environments including \textit{Montezuma’s Revenge}.
The LGE implementation is available as open-source at \url{https://github.com/qgallouedec/lge}.
\end{abstract}

\section{Introduction}

RL algorithms aim to learn a policy by maximizing a reward signal. In some cases, the rewards from the environment are sufficiently informative for the agent to learn a complex policy, and therefore achieve impressive results, including world level in Go \cite{silver2016mastering}, StarCraft \cite{vinyals2019grandmaster}, or learning sophisticated robotic tasks \cite{lee2019robust}.
However, many real-world environments provide extremely sparse \cite{bellemare2016unifying}, deceptive \cite{lehman2011abandoning} rewards, or none at all.
In such environments, random exploration, on which many current RL approaches rely, may not be sufficient to collect data that is diverse and informative enough for the agent to learn anything.
In these cases, the agent must adopt an efficient exploration strategy to reach high reward areas, which may require a significant amount of interactions.

Recently, \citet{ecoffet2021first} introduced a new paradigm in which a goal-conditioned agent is trained to reach states it has already encountered, and then explore from there. The agent thus iteratively pushes back the frontier of its knowledge of the environment. We call this family of algorithms \textit{return-then-explore}. \citet{ecoffet2021first} provide Go-Explore, an algorithm of this new family, that outperforms by several orders of magnitude the state-of-the-art scores on the game \textit{Montezuma's Revenge}, known as a hard-exploration problem.
Go-Explore relies on a grouping of observations into \textit{cells}. These cells are used both to  select target observations at the frontier of yet undiscovered states and to build a subgoal trajectory for the agent to follow to reach the final goal cell. As \citet{ecoffet2021first} initially spotted, the cell design is not obvious. It requires detailed knowledge of the observation space, the dynamics of the environment, and the subsequent task. If any important information about the dynamics of the environment is missing from the cell representation, the agent may fail to explore at all. For example, in \textit{Montezuma's Revenge}, possession of a key is a crucial piece of information that when included in the cell representation increases exploration by several orders of magnitude.
We also demonstrate in Appendix \ref{appendix:cell_criticality} that the cell design has a major influence on the results.

In this paper, we present Latent Go-Explore (LGE), a new algorithm derived from Go-Explore which operates without cells. This new algorithm meets the definition of a \textit{return-then-explore} family of algorithms since the agent samples a final goal state at the frontier of the achieved states, returns to it, and then explores further from it. 
Our main contribution consists of three major improvements.

\begin{figure*}
    \centering
    \includegraphics[width=0.9\textwidth]{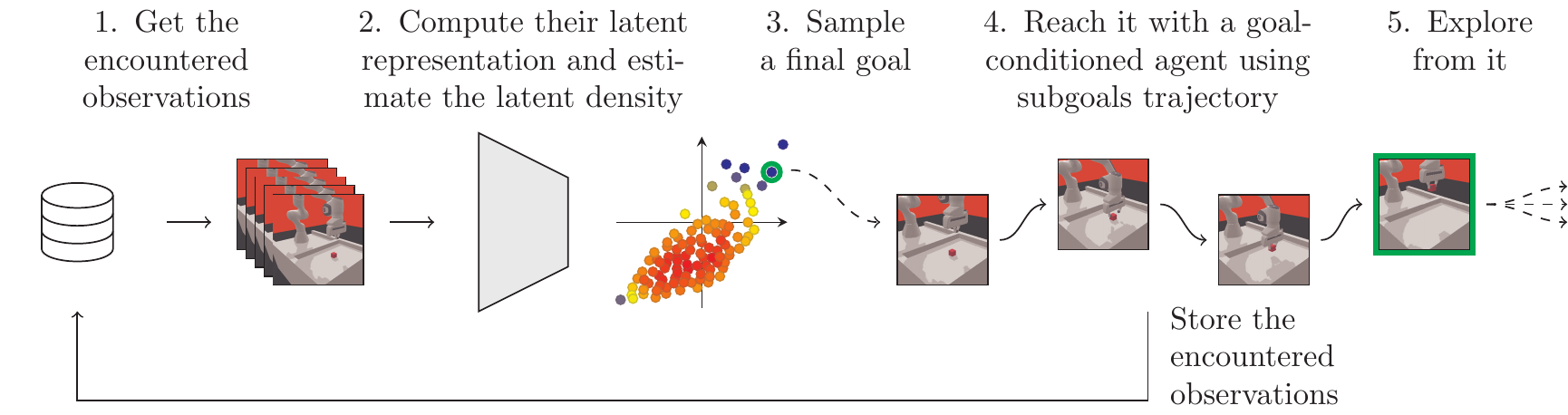}
    \caption{LGE exploration workflow. The encountered observations are encoded in a latent space. A latent density is estimated. A final goal is sampled from the states already reached, by skewing the distribution with the density. A goal-conditioned agent is trained to reach this goal by pursuing a sequence of subgoals, derived from the experiment that led to the final goal. Once the agent has reached the final goal, it explores from it with any exploration strategy.}
    \label{fig:main_figure}
\end{figure*}

\begin{itemize}
    \item A latent representation is learned simultaneously with the exploration of the agent to provide the most up-to-date and informative representation possible.
    \item Sampling of the final goal is based on a non-parametric density model in latent space. This leverages the learned latent representation for sampling the states of interest to be reached.
    \item The subgoal path pursued by the agent is reduced using a characteristic latent distance.
\end{itemize}

These three modifications, detailed in Section \ref{sec:lge}, allow us to generalize the Go-Explore approach to any continuous high-dimensional environment. It also enables the automation of the encoding of observations into an informative latent representation, eliminating the need for manual cell design. The full LGE exploration workflow is presented in Figure \ref{fig:main_figure}.

To evaluate LGE, we conducted experiments in the context of reward-free exploration in various hard-exploration environments including a maze, a robotic arm interacting with an object, and two Atari games known for their high exploration difficulty: \textit{Montezuma's Revenge} and \textit{Pitfall}.

LGE can use various types of latent representation learning methods. In this study, we demonstrate the use of three such methods, including inverse dynamics, forward dynamics, and auto-encoding mechanism.
We show in Section \ref{subsec:results} that for the environments studied, LGE outperforms all state-of-the-art algorithms studied in this paper, and in particular Go-Explore for the exploration task.

\section{Preliminaries and Related Work}

\subsection{Preliminaries}

\paragraph{Markow Decision Process}
This paper uses the standard formalism of a discounted Markov Decision Process (MDP) defined as the tuple $(\mathcal{S},\mathcal{A}, \mathcal{P}, \mathcal{R}, \gamma, \rho_0)$ where $\mathcal{S}$ is the set of states, $\mathcal{A}$ is the set of actions, $\mathcal{P} : \mathcal{S} \times \mathcal{A} \to \mathcal{S} $ is the (unknown) transition function providing the probability distribution of the next state given a current state and action, $\mathcal{R} : \mathcal{S} \times \mathcal{A} \times \mathcal{S} \to \mathbb{R}$ is the reward function, $\gamma$ is the discount factor and $\rho_0$ is the initial distribution of states.
A policy, denoted $\pi:\mathcal{S}\times \mathcal{A}\to \mathbb{R}^+$ is the probability distribution such that $\pi(a|s)$ is the probability of choosing action $a$ in state $s$. We denote the previously defined values with discrete time $t$ such that $s_t$, $a_t$ and $r_t$ denote respectively the state, action, and reward at timestep $t$. 
The goal is to learn a policy $\pi$ that maximizes the long-term expected reward $\mathbb{E}_{}[\sum_{t=0}^{+\infty}\gamma^t r_t]$. 

\paragraph{Goal-conditioned MDP}
We note that every MDP can be augmented into a goal-conditioned MDP with a goal space $\mathcal{G}$ and an initial goal distribution $\rho_g$. At each timestep, the observation is augmented with a goal and the reward function depends on this goal. A goal-conditioned policy \cite{kaelbling1993learning}, denoted $\pi(\cdot|\cdot, \cdot)$ also depends on the goal. 

\subsection{Related Work}
\label{subsec:related_work}

Exploration in RL can be divided into three types \cite{ladosz2022exploration}: unstructured exploration\footnote{We replace the terminology of \citet{ladosz2022exploration} \textit{random exploration} by \textit{unstructured exploration} that we think is more accurate.}, intrinsic rewards-based methods, and goal-based methods.

\paragraph{Unstructured exploration}

In unstructured exploration, the agent does not adhere to a predetermined exploration plan and instead takes actions randomly or according to a simple heuristic. These actions may be sampled uniformly from the action space, or in the continuous case, they may be augmented by exploration noise that is parametrized by the current state \cite{haarnoja2018soft} or not \cite{lillicrap2016continuous}. Unstructured exploration can be effective in some environments, but it may not be sufficient to explore more complex or sparsely rewarded environments.

\paragraph{Intrinsic rewards-based methods}

Intrinsic rewards-based methods are inspired by the concept of intrinsic motivation in cognitive science \cite{oudeyer2009intrinsic}. They involve the addition of an additional reward signal, called \textit{intrinsic}, to the reward signal from the environment, called \textit{extrinsic}. This intrinsic reward is designed to encourage exploration. It can be based on the state visitation count \cite{bellemare2016unifying, machado2020count} or on the prediction error of a model learned from the collected data \cite{houthooft2016vime, pathak2017curiosity, achiam2017surprise, pathak2019self, burda2019exploration, tao2020novelty}.

\begin{algorithm}[tb]
    \caption{LGE}
    \label{alg:lge}
    \begin{algorithmic}
        \STATE {\bfseries Input:} Number of iterations in the exploration phase $T$.
        \STATE {\bfseries Initialize:} Replay buffer $D=\emptyset$; Encoding module; Goal-conditioned policy $\pi$
        \WHILE{$t<T$}
            \STATE Sample a final goal state with Equation \eqref{eq:sample_goal}.
            \STATE Build the subgoal trajectory $\tau^g$ using Equation  \eqref{eq:lighten_goal_traj}.
            \STATE Initialize the subgoal index: $i \gets 0$.
            \WHILE{the last goal of $\tau^g$ is not reached}
                \STATE Collect interaction using $\pi(\cdot | \cdot, \tau_i^g)$ and store it into dataset $D$.
                \IF {subgoal $\tau^g_i$ is reached, i.e. $||\phi(s_t) - \phi(\tau^g_i) || < d$}
                    \STATE Move to the next subgoal: $i \gets i+1$.
                \ENDIF
            \ENDWHILE
            \STATE Explore until the end of the episode with any exploration strategy.
            \STATE Update $\pi$ with any off-policy algorithm and HER.
            \STATE Every $\mathtt{update\_encoder\_freq}$ timesteps, update encoder $\phi$ with any representation learning algorithm.
        \ENDWHILE
        \STATE \textbf{return} the goal-conditioned policy $\pi$ and the dataset $D$.
    \end{algorithmic}
\end{algorithm}

\paragraph{Goal-based methods}

Methods that modify the reward of the environment provide no mechanism for distilling the knowledge gained from visiting various states. Agents may visit new states, but they quickly forget about them when other states become newer. To address this issue, recent work has suggested the use of a goal-conditioned autotelic agent specifically trained for the exploration task. This approach allows for the use of the knowledge gained during exploration to realize new user-specified goals \cite{levine2021understanding, colas2022autotelic}. During the exploration phase, the reward signal is ignored, and after the exploration phase, the data collected by the agent is used to learn one or more subsequent tasks \cite{jin2020reward}. Goal-based methods condition the agent with a goal that is used to guide exploration towards unknown areas. These methods rely on a goal generator to create goals for the agent. We divide goal-based methods into two categories: \textit{exploratory goal} methods and \textit{goals to explore from} methods (called \textit{post-exploration} in \cite{yang2022first}).

\textit{Exploratory goal} methods follow the intuition that the agent discovers new areas of the observation space by pursuing goals that have been little or not achieved before. 
The challenge of these methods is to choose the goal to be neither too easy nor too hard. The literature contains several ways to approach this trade-off. Some methods sample goals that either maximize Learning Progress \cite{colas2019curious, portelas2020teacher} or value disagreement \cite{zhang2020automatic}. Other methods sample goals from the least visited areas using a parametric density model on the visited states \cite{pong2020skew}. It is also possible to imagine goals that have never been reached using a language model \cite{colas2020language}, a generative model \cite{racaniere2020automated} or a GAN \cite{florensa2018automatic}. 

In \textit{goals to explore from} methods the agent samples a goal from previously visited states. It returns to it, either by teleportation \cite{ecoffet2019go, matheron2020efficient}, or using a goal-conditioned policy \cite{ecoffet2021first}. The challenge of these methods is to choose a goal that is of high exploratory interest. Similarly, some methods estimate the density of the encountered states, using either parametric methods \cite{pitis2020maximum} or non-parametric methods \cite{ecoffet2021first, matheron2020efficient}, to target the low-density areas.

In summary, methods based on a goal reaching policy should facilitate scalable RL. The stunning results of Go-Explore illustrate this point but remain circumscribed to few environments and require a lot of domain knowledge to work. By bridging with concepts already used in the intrinsic reward literature, we show a way to make this approach more general and simpler.

\section{Latent Go-Explore}
\label{sec:lge}

LGE meets the definition of the \textit{return-then-explore} family of algorithms. First, a final goal state is sampled from the replay buffer, then the agent learns a goal-conditioned policy to reach this goal. When the agent reaches the goal, the agent starts to explore.
LGE learns a latent representation of observations and samples the goal pursued by the goal-conditioned agent in priority in low latent density areas. In Section \ref{subsec:learning_latent}, we present how the latent representation of observations is learned. In Section \ref{subsec:density_estimation_for_intrinsic_goal_sampling}, we show how the latent density is estimated and how the final goal state pursued by the agent is sampled. Finally, in Section \ref{subsec: subgoal_trajectory}, we show how to build a subgoal trajectory from the final goal to increase the agent's performance, in particular in far-away goal situations.
The pseudo-code of the resulting algorithm is presented in Algorithm \ref{alg:lge}.

\subsection{Learning a latent representation}
\label{subsec:learning_latent}



The literature contains several latent representation learning methods for RL. Learning such a representation is orthogonal to our approach. Hence, LGE can be combined with any learning method without the need for further modifications. Choosing the best representation learning method given the environment is out of the scope of this paper.
In this paper, we present three methods of representation learning that have been found to work well with our test environments. Two of these methods are inspired by the literature on intrinsic reward-based methods,

\paragraph{Inverse dynamic representation learning}
\citet{pathak2017curiosity} proposed an intrinsic reward calculated based on the agent's prediction error of the consequence of its own actions. The representation is learned using two submodules. The first encodes the observation into a latent representation $\phi(s_t)$. The second takes as input $\phi(s_t)$ and $\phi(s_t+1)$ and outputs the prediction of the action taken by the agent at time step $t$. The parameters $\theta$ of the inverse model $\mathcal{P}^\mathrm{inv}_\theta$ are optimized by minimizing the loss function:

\begin{equation}
    L = \frac{1}{|N|} \sum_{(s_t, a_t, s_{t+1}) \sim D} \frac{1}{2} \lVert a_t - \mathcal{P}^\mathrm{inv}_\theta(s_t, s_{t+1})\rVert^2_2  
\end{equation}

The inverse dynamics representation learning allows getting a latent representation of the states containing only the aspects of the state on which the agent can have an influence.

\paragraph{Forward dynamic representation learning}
In \citet{achiam2017surprise}, the intrinsic reward is calculated based on the prediction error of a model approximating the transition probability function of the MDP.
Two submodules are used. The first one encodes the observation to a latent representation $\phi(s_t)$. The second takes as input $\phi(s_t)$ and $a_t$ and outputs the prediction of the next state $\hat{s}_{t+1}$.
The model parameters $\theta$ are optimized by minimizing the loss function:

\begin{equation}
    L= -\frac{1}{|N|} \sum _{(s_t, a_t, s_{t+1}) \sim D} \log\mathcal{P}_\theta(s_{t+1} \mid s_t, a_t)
\end{equation}

\paragraph{Vector Quantized Variational Autoencoder (VQ-VAE)}

Autoencoding \cite{hinton2006reducing} aims to train a neural network to reconstruct its input by learning a compressed representation of the data. This approach is known to be effective in extracting useful features from the input, especially images. For Atari environments, we use a VQ-VAE \cite{oord2017neural}, a technique that combines autoencoding with vector quantization, and has shown good results, while being simple to train. We use the coordinates of the embeddings in the embeddings table as the latent representation.

\subsection{Density estimation for intrinsic goal sampling}
\label{subsec:density_estimation_for_intrinsic_goal_sampling}

The success of the proposed method relies on the agent's ability to generate for itself goals that it will be able to reach and then explore from there. For the agent to progress in the exploration of the environment, these goals must be at the edge of the yet unexplored areas. To identify these areas, we use an estimator of the density of latent representations of the encountered states. Moreover, we require the goal to be reachable. The set of reachable states is a subset of the state space that we assume to be unknown. The easiest way to satisfy the previous requirement is therefore to sample among the states that have already been reached.

We estimate the density of latent representations of the encountered states (called latent density) using the particle-based entropy estimator originally proposed by \cite{kung2012optimal} and used in the literature on intrinsically motivated RL \cite{liu2021behavior, liu2021active}. This estimator has the advantage of being nonparametric and thus does not hinge on the learning capabilities of a learned model.
Appendix \ref{appendix:density_estimation} describes the details of the implementation of this estimator, denoted $\hat{f}$.

The sampling of the final goal state follows a geometric law on the rank in the latent density sort $R_{i}$. The probability to draw $s_i$ as the final goal state is

\begin{equation}
    \mathbb{P}(G=s_i)=(1-p)^{R_{i}-1}p
    \label{eq:sample_goal}
\end{equation}

where $G$ is the random variable corresponding to the final goal state, and $0\leqslant p\leqslant 1$ is a hyperparameter controlling the bias towards states with a low latent density.

This method has the advantage of being robust to approximation errors in the density evaluation, which can be particularly important in low density areas. In doing so, we only focus on the ability of the model to correctly order the observations according to their latent density.

The representation is jointly learned with the exploration of the agent. Therefore, the latent density must be regularly recomputed to take into account the most recent representation on the one hand, and the recently visited states on the other hand. However, considering the slow evolution of this value, we choose to recompute the latent density only once every 5k timesteps for maze and robotic environments, and every 500k timesteps for Atari environments. This allows us to significantly reduce the computation needs while having a low empirical impact on the results.

\subsection{Subgoal trajectory}
\label{subsec: subgoal_trajectory}

As learning progresses, the sampled final goal states are increasingly distant. However, reaching a distant goal is challenging because it implies a sparse reward problem.

To overcome this problem, we condition the agent to successive intermediate goals $\tau^g = (g_0, g_1 ..., g_L)$ that should guide it to the final goal state $g_L$. These intermediate goals are chosen from the trajectory that led the agent to the final goal state $(s_0, s_1, \ldots, s_T)$. 

The trajectory that led the agent to the final goal state is unlikely to be optimal. Plus, if the agent is conditioned by the whole trajectory, it may fail to reach all of them, even though some of them may not be necessary to reach the final goal state. To allow the agent to find a better path to the final goal state, we remove some subgoals from this trajectory. To decide whether a subgoal should be removed from the trajectory, we evaluate the latent distance to the previous subgoal. If the distance is less than the threshold, then the goal is removed.

\begin{equation}
    \forall i \leqslant L - 1,\quad ||\phi(g_i) - \phi(g_{i+1})|| > d
    \label{eq:lighten_goal_traj}
\end{equation}

Unlike Go-Explore, LGE don't use the best known trajectory that leads to the sampled goal area (cell). The main reason is that the best known trajectory may be particularly difficult to reproduce, due to the dynamics and the stochasticity of the environment, or cause the early termination of the episode.
For example, in \textit{Montezuma's Revenge}, there are two ways to reach the bottom of the left ladder (a necessary step to get the first key). The first one consists in jumping right from the promontory, but causes the death of Panama Joe (the character) and as a result ends of the episode. The second one, longer, consists in going around by the right ladder. 
Therefore, if the agent always chooses the shortest path (like Go-Explore), it will most likely fail to reach the first key and to further explore the environment.

Once the goal is reached, the agent explores using any exploration strategy. For the sake of simplicity, we choose a random exploration strategy for our experiments. We also impose that the agent repeats the previous action with a probability of 90\%. This technique has been shown to increase the results significantly \cite{ecoffet2021first}.

\section{Experiments}

To demonstrate the effectiveness of our method, we apply it to a range of pure exploration tasks. We focus on environments for which naive random exploration is not sufficient to explore the rich variety of reachable states.
We compare the results obtained with LGE with the results obtained using several algorithms based on intrinsic curiosity and others based on goal-directed strategies. 
For each environment, LGE uses the representation method that empirically gives the best results. Consequently, we use the forward dynamics for the maze environment, the inverse dynamics for the robotic environment, and the VQ-VAE for Atari.

In terms of infrastructure, each run was performed on a single worker machine equipped with one CPU and one NVIDIA\textregistered~ V100 GPU + 120 Gb of RAM.

\subsection{Environments}

\paragraph{Continuous maze}
We train an agent to navigate in a continuous 2D maze. The corresponding configuration is shown in Figure \ref{fig:scene_coverage_illustration}. The agent starts every episode in the center of the maze. At each timestep, the agent receives the current coordinates as an observation and chooses an action that controls its location change. If the agent collides with a wall, it returns to its previous position. The reachable space is a square of $12 \times 12$ and the agent's action is limited to $[-1, 1]$ horizontally and vertically. The agent can interact 100 times with the environment (which is just enough to explore all the maze), after which the episode ends.

\paragraph{Robotic environment}
Robotic environments are interesting and challenging application cases of RL, especially since the reward is often sparse. We simulate a Franka robot under the PyBullet physics engine using panda-gym  \cite{gallouedec2021pandagym}. The robot can move and interact with an object. 
The agent has access to the position of the end-effector and the position of the object, as well as to the opening of the gripper. The agent interacts 50 times with the environment and then the object and the robot arm are reset to their initial position.

\paragraph{Atari}

We train LGE on two high-dimensional Atari 2600 environments simulated through the Arcade Learning Environment (ALE, \citet{bellemare2013arcade}) that are known to be particularly challenging for exploration: \textit{Montezuma's Revenge} and \textit{Pitfall}. Details of the settings used are presented in Appendix \ref{appendix:hyperparam}.

\subsection{Baselines}

\subsubsection{Random exploration}
\label{subsec:baselines}
Most RL methods from the literature do not follow any structured exploration strategy. In a reward-free context, the performance of the latter is often equivalent to a random walk. We take as a reference a random agent, whose actions are uniformly sampled over the action space at each time step, Soft Actor-Critic (SAC, \citet{haarnoja2018soft}) and Deep Deterministic Policy Gradient (DDPG, \citet{lillicrap2016continuous}) for continuous action space environments.

\begin{figure*}
    \centering
    \begin{subfigure}[b]{0.15\textwidth}
        \centering
        \includegraphics[width=\textwidth]{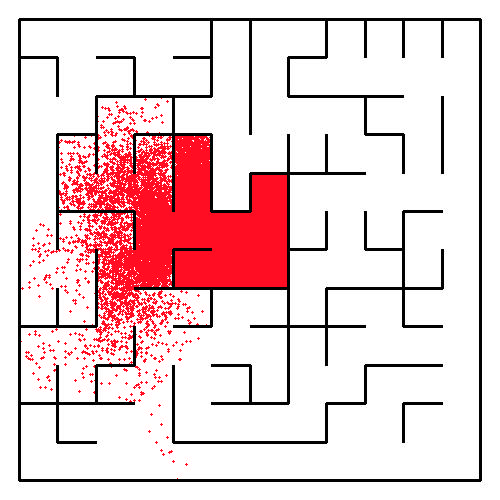}
        \caption{Random}
        \label{fig:random_100k}
    \end{subfigure}
    \begin{subfigure}[b]{0.15\textwidth}   
        \centering   
        \includegraphics[width=\textwidth]{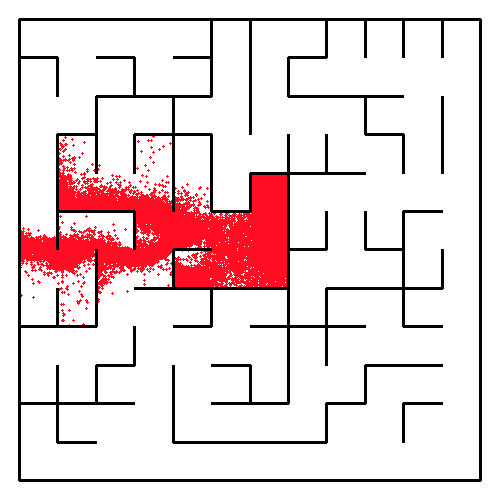}   
        \caption{SAC}   
        \label{fig:sac_100k}
    \end{subfigure}
    \begin{subfigure}[b]{0.15\textwidth}   
        \centering   
        \includegraphics[width=\textwidth]{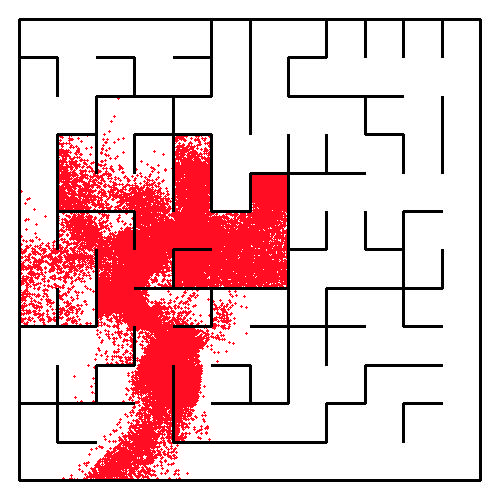}   
        \caption{SAC+ICM}   
        \label{fig:icm_100k}
    \end{subfigure}
    \begin{subfigure}[b]{0.15\textwidth}
        \centering
        \includegraphics[width=\textwidth]{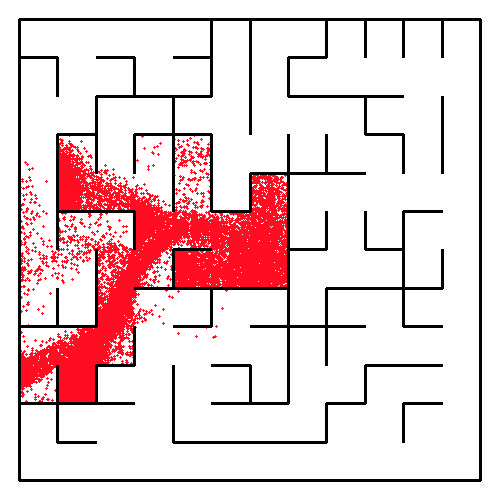}
        \caption{SAC+Surprise}
        \label{fig:surprise_100k}
    \end{subfigure}

    \begin{subfigure}[b]{0.15\textwidth}
        \centering
        \includegraphics[width=\textwidth]{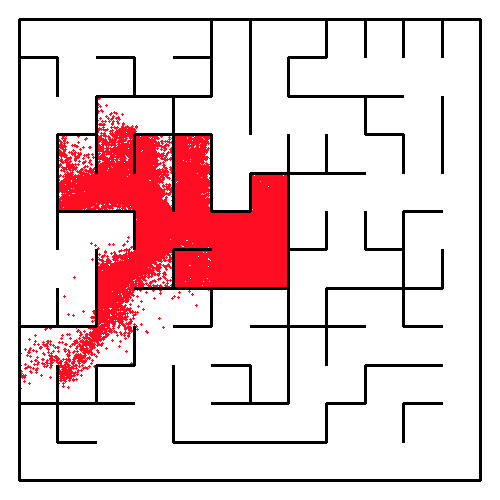}
        \caption{Skew-Fit}
        \label{fig:skewfit_100k}
    \end{subfigure}
    \begin{subfigure}[b]{0.15\textwidth}   
        \centering   
        \includegraphics[width=\textwidth]{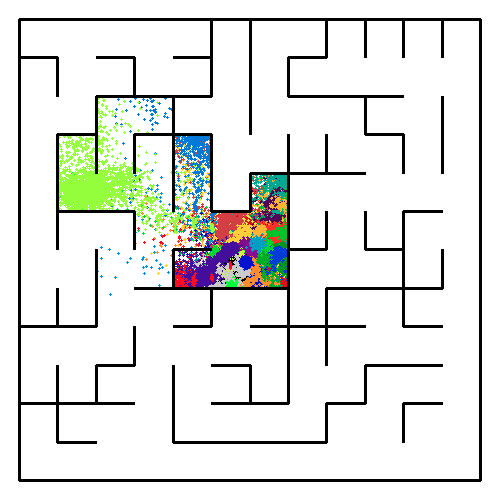}   
        \caption{DIAYN}   
        \label{fig:diayn_100k}
    \end{subfigure}
    \begin{subfigure}[b]{0.15\textwidth}
        \centering
        \includegraphics[width=\textwidth]{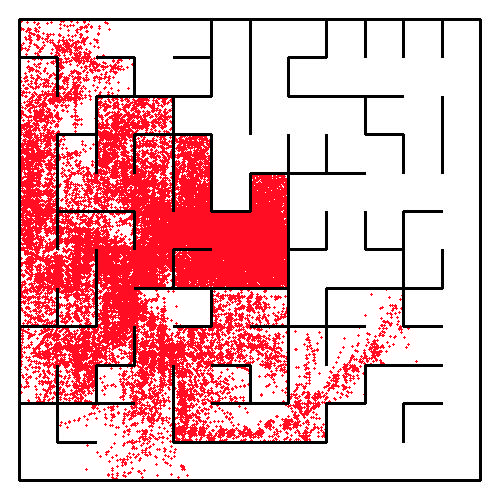}
        \caption{Go-Explore}
        \label{fig:go_explore_100k}
    \end{subfigure}
    \begin{subfigure}[b]{0.15\textwidth}
        \centering
        \includegraphics[width=\textwidth]{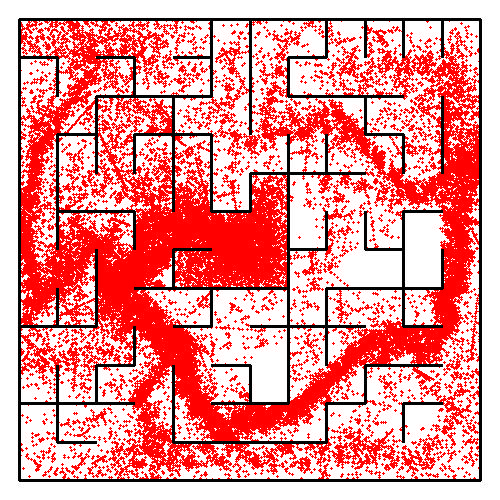}
        \caption{LGE (ours)}
        \label{fig:lge_100k}
    \end{subfigure}
    
    \caption{Space coverage of the maze environment after 100k timesteps. In \subref{fig:diayn_100k}, the different colors are the different skills.}
    \label{fig:scene_coverage_illustration}
\end{figure*}

\subsubsection{Intrinsic reward-based exploration}

In this paper, we take as reference two widely used intrinsic reward-based methods combined with either SAC or DDPG. These methods stand out from the others because, despite their simplicity, they have demonstrated good performance on a wide variety of tasks.

\paragraph{Intrinsic Curiosity Module} (ICM, \citet{pathak2017curiosity}) The intrinsic reward is computed as the mean square error between the true latent representation and the one predicted by a learned dynamic model given the action taken. The encoder is trained jointly with an inverse dynamics model.
\paragraph{Surprise} \cite{achiam2017surprise} The intrinsic reward is the approximation of the KL divergence between the actual transition probabilities and a learned transition model.

\subsubsection{Goal-directed exploration}

\paragraph{Go-Explore} \label{subsec:baselines-go-explore} 
The agent divides the observation space into cells, prioritizes the cells that have been visited the least, returns to them using a goal-conditioned policy, and then continues exploring from that point. This is the policy-based and without domain knowledge variant of Go-Explore, but we refer to it simply as \textit{Go-Explore}.
The observations in continuous environments are converted into cell representations by discretizing them. In the maze environment, we use a $24 \times 24$ grid, and in the robotic environment, we use a grid with a 0.1m resolution for the position of the gripper and object.
For Atari, we use the same fixed cell representation as proposed by \cite{ecoffet2021first} in the policy-based case: the observation is converted to grayscale and reduced to the size of $8\times 11$ pixels. The depth is then reduced from 256 to 8 values according to $\lfloor\frac{8p}{255}\rfloor$ where $p$ is the pixel value. The resulting image is the representation of the cell.
Go-Explore is the closest baseline to our algorithm. Appendix \ref{appendix:ge_vs_lge} details the differences between LGE and Go-Explore.

\paragraph{Diversity Is All You Need} (DIAYN, \citet{eysenbach2019diversity}) The agent is conditioned by a \textit{skill} and a discriminator predicts the skill pursued by the agent. The more the discriminator predicts with certainty the skill pursued, the bigger the reward. Conjointly, the discriminator is trained to maximize the distinguishability of skills.

\paragraph{Skew-Fit} \cite{pong2020skew} The agent's goal sampling is skewed to maximize the entropy of a density model learned on the achieved states.

For the goal-directed methods, we use Hindsight Experience Replay (HER, \citet{andrychowicz2017hindsight}) relabeling which has shown to significantly increase learning.

To nullify the variation in results due to different implementations, we implement all algorithms in the same framework: Stable-Baselines3 \cite{raffin2021stable}.
The set of intrinsic reward-based methods and goal-directed methods are underpinned by the same off-policy algorithm. The hyperparameters for this algorithm are identical. 
For the maze environment, we use SAC, while for the robotic environment, we use DDPG as it gives better results for all methods. For Atari environments, we use QR-DQN \cite{dabney2018distributional}, as it commonly considered to be a strong baseline on it.
For Atari, we only compare LGE to Go-Explore as it far outperforms the others.
To negate the influence of a bad choice of hyperparameter on the results, the method-specific hyperparameters are optimized. Appendix \ref{appendix:hyperparam} details the optimization process for all methods and the resulting hyperparameters.

\subsection{Measuring the exploration}

In this paper, we focus on the agent's ability to explore its environment in a pure exploration context, i.e. in the absence of extrinsic reward. This step is particularly important because, in the case of an environment with very sparse rewards, the agent can interact a large number of times with the environment without getting any reward. It is therefore necessary to follow an efficient exploration strategy to discover the few areas of the state space where the agent can get a reward.
To be able to compare the results obtained by different methods in this context, it is necessary to use a common metric for the quality of exploration.

\begin{figure}
    \centering
    \includegraphics[width=\columnwidth]{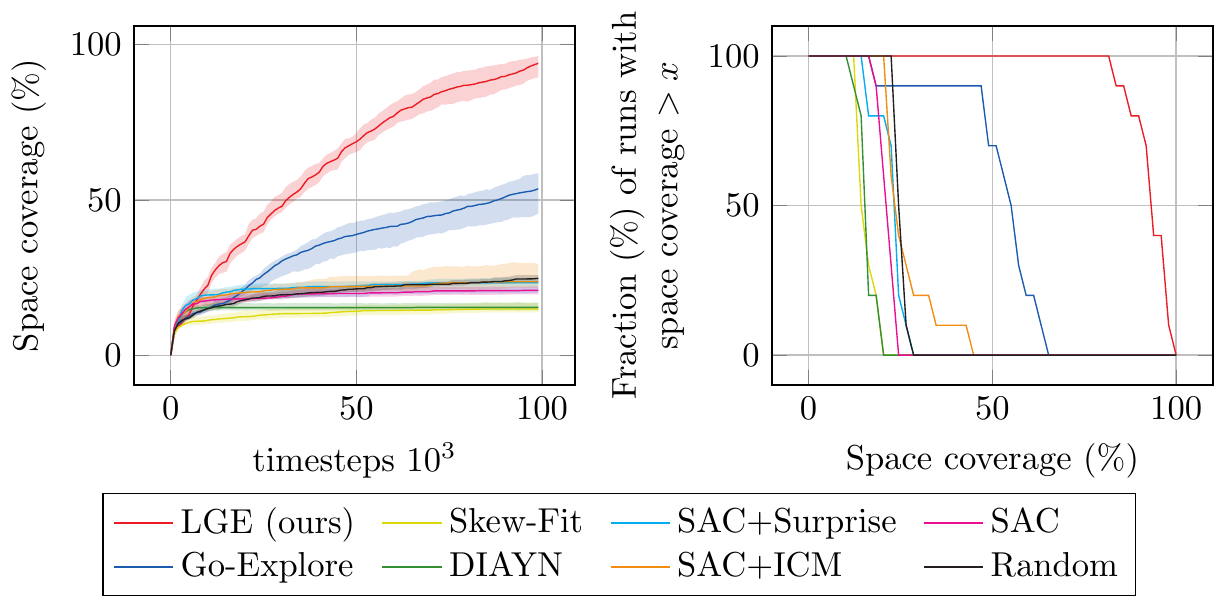}
    \caption{Comparison of the space coverage of the maze environment. Each experiment is run 10 times. The left plot represents the space coverage (number of cell divided by the total number of reachable cells) across timesteps. The solid lines are the IQMs and the shaded areas are the 95\% confidence intervals. The right plot is the final performance profile (higher is better).}
    \label{fig:maze_curves}
\end{figure}

The literature uses various metrics. Some papers use the average reward on a hard-exploration task \cite{taiga2020on}, the zero-shot performance on a predefined task \cite{sekar2020planning} or monitor specific identifiable events in the environment that indirectly informs the degree of exploration  \cite{gulcehre2020making}.
We argue that these indirect measures are unsatisfactory as they rely on the subsequent learning ability of an online and offline agent respectively.
For simplicity, we use the number of visited cells as the metric, whose construction strategy is explained in Section \ref{subsec:baselines-go-explore}. Therefore, the figures represent the number of cells explored, although Go-Explore is the only algorithm to explicitly maximize this metric.
Following the guidelines of \cite{agarwal2021deep}, we use for all plots in this paper the interquartile mean (IQM) with the 95\% confidence interval. 

\subsection{Main Results}
\label{subsec:results}

The exploration results for the maze environment are presented in Figure \ref{fig:maze_curves}. A rendering of the positions explored by the agent is presented in Figure \ref{fig:scene_coverage_illustration}. We note that only LGE and Go-Explore significantly outperform the results obtained with random exploration. This demonstrates the effectiveness of the \textit{return-then-explore} paradigm in this environment. We note that exploration based on intrinsic curiosity does not yield significantly better results than those obtained by random exploration. We hypothesize that the simple dynamics of the environment makes the intrinsic reward to quickly converge to $0.0$. Surprisingly, neither Skew-Fit nor DIAYN performs significantly better than random exploration. For DIAYN, we find that most of the skills were concentrated in the initial position area of the agent. We hypothesize that this is the consequence of the lack of \textit{post-exploration} described by \cite{yang2022first}. Finally, we note that, although the cell size has been optimized, LGE significantly outperforms Go-Explore. LGE manages to cover almost the entire reachable space at the end of the runs while exhibiting low variability in the results. 

\begin{figure}
    \centering
    \includegraphics[width=\columnwidth]{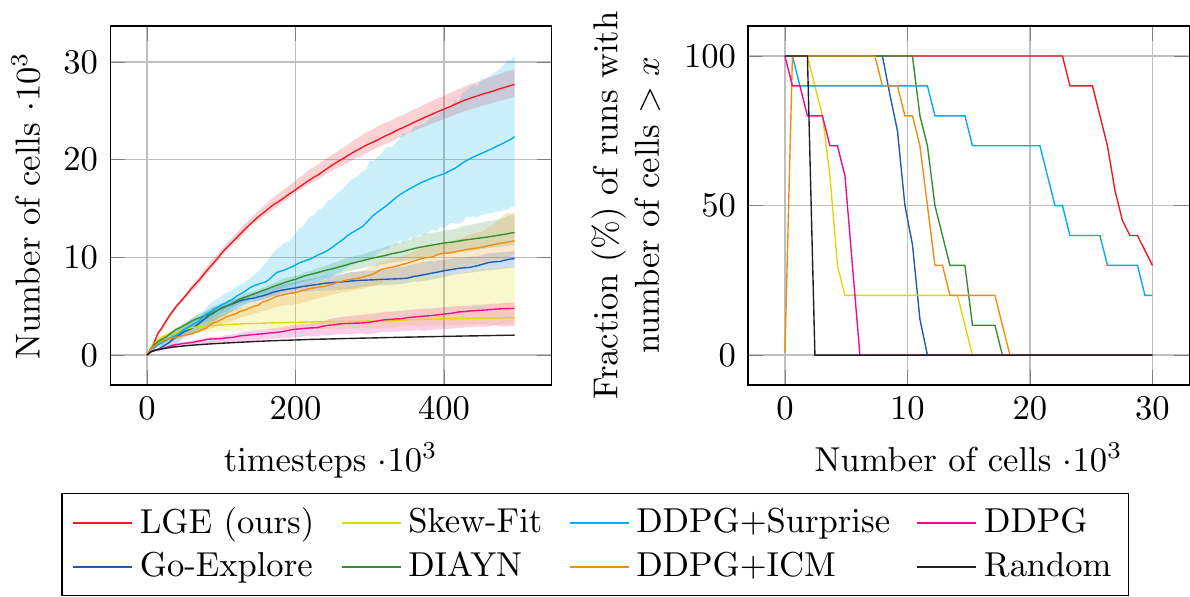}
    \caption{Comparison of exploration with the robotic environment. Each experiment is run 10 times. The left plot represents the number of explored bins across timesteps. The solid lines are the IQMs and the shaded areas are the 95\% confidence intervals. The right plot is the final performance profile (higher is better).}
    \label{fig:panda_curves}
\end{figure}

The exploration results for the robotic environment are presented in Figure \ref{fig:panda_curves}. We notice that LGE significantly outperforms all other methods. Notably, Go-Explore performs only slightly better than random exploration. We note that Go-Explore does not learn to grasp the object throughout the learning process. The results presented on a robotic environment by \citet{ecoffet2021first} are much better. We presume this is mainly due to the meticulous work done on the state space examination and the induced cell design. Here, we use a naive grid-like cell design. Although the grid parameter is optimized, it does not yield good exploration results with this environment. We thus demonstrate the benefit of using a learned representation to automatically capture important features of the environment's dynamics. 

\begin{figure}[b!]
    \centering
    \includegraphics[width=\columnwidth]{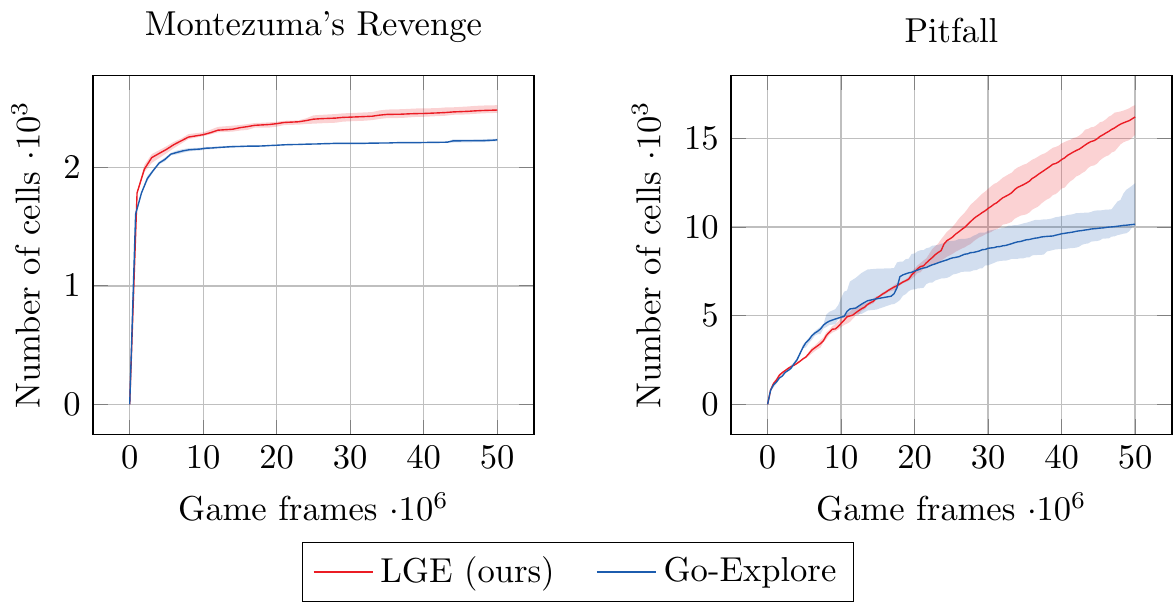}
    \caption{Comparison of exploration on the Atari environments. Each experiment is run 3 times. The solid lines are the IQMs and the shaded areas are the 95\% confidence intervals.}
    \label{fig:atari_curves}
\end{figure}

The exploration results for Atari are presented Figure \ref{fig:atari_curves}. We see that both LGE and Go-Explore quickly discover a large number of cells, then continue their exploration by regularly discovering new cells. LGE slightly outperforms Go-Explore on both \textit{Pitfall} and \textit{Montezuma's Revenge}.
Nevertheless, we note that the number of discovered cells is much lower than that of Go-Explore in its full configuration (around 5k for \textit{Montezuma's Revenge}), including domain knowledge and the ability to reset the environment in any state. This shows the criticality of these settings for exploring these particular environments.

\subsection{Ablation study}

\begin{figure}
    \centering
    \includegraphics[width=\columnwidth]{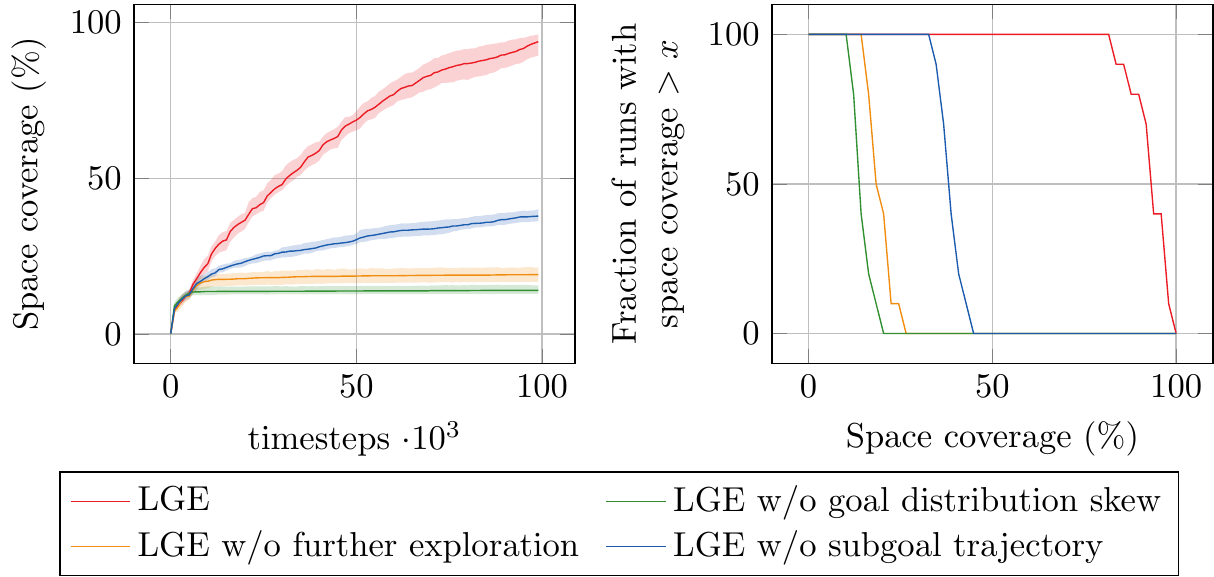}
    \caption{Result of ablation study on the maze environment. Each experiment is run 10 times. The left plot represents the space coverage across timesteps. The solid lines are the IQMs and the shaded areas are the 95\% confidence intervals. The right plot is the final performance profile (higher is better).}
    \label{fig:ablation_study}
\end{figure}

We study the impact of the ablation of three key elements of LGE.
(1) LGE without further exploration (as exploratory goal methods, see Section \ref{subsec:related_work}): the environment is reset once the agent reaches the final goal instead of performing the exploration random interactions.
(2) LGE without skewing the final goal distribution in favor of low latent density areas, the final goals are sampled uniformly among the reached goals.
(3) LGE without subgoals trajectory reduction: the agent is just conditioned by the final goal state.
We perform these ablation studies on the maze environment and use the same hyperparameters as in Section \ref{subsec:results}. The results are shown in Figure \ref{fig:ablation_study}.



Our ablation study reveals that the three ablations have a significant impact on the results. We find that exploration after reaching the final goal is crucial, confirming the results of \cite{yang2022first}; without it, the agent reaches the limits of its knowledge but has little chance to explore further. Additionally, sampling goals with low latent density can significantly improve results by directing exploration to states with high exploratory value. Furthermore, we observe that conditioning the agent with successive subgoals greatly improve its exploration.

\section{Discussion}

\subsection{Limitation and future work}

\paragraph{Goal-achievement functions}

In LGE, an agent is considered to have reached a goal (whether final or intermediate) when the latent distance between its state and the goal is below a threshold. This is a naive way of defining a goal achievement function \cite{colas2022autotelic} that depends crucially on the latent representation. We believe that the results could be improved by envisioning a more informative and suitable goal achievement function for our method. 

\paragraph{The initial state must remain the same across episodes} 

The approach we propose is based on the assumption that the agent is always initialized in the same state. This assumption guarantees that at the beginning of each episode, all the states previously reached are reachable and that the subgoal trajectory starts with the initial state of the agent.
However, in some environments, especially in procedurally generated environments, this assumption is not fulfilled \cite{kuttler2020nethack}. 
In this situation, trying to follow the subgoal trajectory may be counterproductive in reaching the final goal. It is also possible that the final goal is not even reachable.
However, note that even the pursuit of an unreachable goal can foster exploration.
We believe that an approach inspired by generative networks such as \cite{racaniere2020automated} may be appropriate to overcome this problem.

\paragraph{High-dimension environments and representation learning} 

Our main contribution consists in the generalization of the Go-Explore approach by using a latent representation.
LGE is notably effective in high-dimensional environments, specifically those with image observations.
Representation learning is the keystone of the method. We provide a proof of concept for a forward model, an inverse model, and a VQ-VAE. We believe that the results can be greatly improved by choosing more finely the representation learning method for each environment by taking advantage of the many works dealing with this subject \cite{lesort2018state}.

The representation used by Search on Replay Buffer (SoRB) \cite{eysenbach2019search} is directly that of the critic. Using the same reward structure as LGE, the critic thus has the nice property of basically learning the negative distance of the shortest directed path between two states.
Overall, we believe that the use of SoRB in the "Go" phase can be a substantial improvement of LGE and is a promising way to solve the three limitations mentioned above. 

Finally, we believe that the community should endeavour to find a relevant metric for exploration, especially for image-based environments. We expect that such a metric would allow a more accurate comparison of different methods.

\subsection{Conclusion}

We introduce LGE, a new exploration method for RL. In this method, our agent explores the environment by selecting its own goals based on a jointly learned latent representation. LGE can be used as pre-training in environments where rewards are sparse or deceptive. Our main contribution is to generalize the Go-Explore algorithm, allowing us to benefit from representation learning algorithms for exploration. 
We present statistically robust empirical results conducted on diverse environments, including robotic systems and Atari games, that demonstrate our approach's significant improvement in exploration performance.

\section*{Acknowledgements}

We would like to thank Adrien Ecoffet, Joost Huizinga, and Jeff Clune for their encouraging feedback and for the time they spent discussing this exciting topic.
We would also like to thank the reviewers of our article for their efforts in reviewing and providing valuable comments. Their comments and suggestions were very helpful in converging on the final version of this work.
This work was granted access to the HPC resources of IDRIS under the allocation 2022-[AD011012172R1] and 2022-[AD011013894] made by GENCI.

\bibliography{biblio}
\bibliographystyle{icml2023}

\newpage
\appendix

\section{Hyperparameters and environments settings}
\label{appendix:hyperparam}

To limit the impact of the large variability of results depending on the hyperparameters, we chose to optimize the hyparameters for each experiment. For maze and robotic environments, we selected 100 unique sets of hyperparameters from a search space presented in Table \ref{table:param_space} using Optuna \cite{akiba2019optuna}. For each hyperparameter set, we train the model with 3 different seeds and keep the median score. For Atari, we selected 10 unique sets of hyperparameters and train the agent just once.

\begin{table*}
\caption{Search space and resulting hyparparameters after optimization.}
\label{table:param_space}
\vskip 0.15in
\begin{center}
\begin{small}
  \begin{tabular}{llllll}
  \toprule
    \textsc{Method} & \textsc{Hyperparameter} & \textsc{Search space} & \textsc{Maze} & \textsc{Robotic} & \textsc{Atari}\\ \midrule
    \textsc{LGE} & Latent distance threshold & $[0.1, 0.2, 0.5, 1.0, 2.0]$ & $1.0$ & $1.0$ & 2.0\\
    & Latent dimension & $[4, 8, 16, 32, 64]$ & $16$ & $8$ & $8\times 8\times 8$ $^c$\\  
    & Geometric parameter & $[0.001, 0.002, 0.005, 0.01, 0.02, 0.05]$ & $0.05$ & $0.01$ & $0.001$\\ \midrule
    \textsc{Go-Explore} & Cell size & $[0.2, 0.5, 1.0, 2.0, 5.0]$ & $2.0$ & $0.2$ & $11\times 8 \times 8$ $^b$ \\ \midrule
    \textsc{ICM} & Scaling factor & $[10^{-3}, 10^{-2}, 10^{-1}, 10^{0}, 10^{1}, 10^{2}]$ & $10^{-1}$ & $10^{-2}$ & N/A\\ 
    & Actor loss coefficient & $[10^{-3}, 10^{-2}, 10^{-1}, 10^{0}, 10^{1}, 10^{2}]$ & $10^{2}$ & $10^{-3}$ & N/A\\ 
    & Inverse loss coefficient$^a$ & $[10^{-3}, 10^{-2}, 10^{-1}, 10^{0}, 10^{1}, 10^{2}]$ & $10^{1}$ & $10^{-3}$ & N/A\\ 
    & Forward loss coefficient$^a$ & $[10^{-3}, 10^{-2}, 10^{-1}, 10^{0}, 10^{1}, 10^{2}]$ & $10^{1}$ & $10^{2}$ & N/A\\
    \midrule
    \textsc{Surprise} & Feature dimension & $[2, 4, 8, 16, 32]$ & 2 & 16 & N/A\\
    & Desired average bonus & $[10^{-2}, 10^{-1}, 10^{0}, 10^{1}]$ & $10^{-2}$ & $10^{-2}$ & N/A\\
    & Model train frequency & $[2, 4, 8, 16, 32, 64, 128]$ & 64 & 8 & N/A\\
    & Model learning rate & $[10^{-6}, 10^{-5}, 10^{-4}, 10^{-3}, 10^{-2}]$ & $10^{-5}$ & $10^{-5}$ & N/A\\ \midrule
    \textsc{DIAYN} & Number of skills & $[4, 8, 16, 32, 64, 128]$ & 32 & 32 & N/A\\ \midrule
    \textsc{Skew-Fit} & Number of models & $[5, 10, 20, 50, 100, 200]$ & 50 & 50 & N/A\\
    & Density power & $[-5.0, -2.0, -1.0, -0.5, -0.2, -0.1]$ & $-1.0$ & $-0.2$ & N/A\\
    & Number of pre-sampled goals & $[64, 128, 256, 512, 1024, 2048]$ & 64 & 128 & N/A\\
    & Success distance threshold & $[0.05, 0.1, 0.2, 0.5, 1.0]$ & 0.5 & 0.2 & N/A\\
\bottomrule
\end{tabular}\\
\end{small}
\end{center}
\footnotesize{$^a$ In the original paper, the sum of the forward and inverse loss coefficients is $1$. We get better results without this constraint.}\\
\footnotesize{$^b$  Width $\times$ Height $\times$ Number of grayscale values.  Not optimized, taken from the original paper.}\\
\footnotesize{$^c$ Width $\times$ Height $\times$ Number of embedding vectors.}
\vskip -0.1in
\end{table*}

The method-specific parameters that have not been optimized are presented in Table \ref{table:specific_hyperparameters}.

\begin{table*}[t]
\caption{Hyperparameters specific to each method. Their value is identical for all experiments and have not been optimized.}
\label{table:specific_hyperparameters}
\vskip 0.15in
\begin{center}
\begin{small}
  \begin{tabular}{lll}
  \toprule
    \textsc{Method} & \textsc{Hyperparameter} & \textsc{Value}\\ \midrule
    \textsc{LGE} & Encoder module trained every $N$ timesteps & $5\mathrm{k}$ (Maze and Robotic), $500\mathrm{k}$ (Atari) \\
    & Learning & $0.001$  \\
    & Batch size & $32$\\
    & Gradient steps & $500$ (Maze and Robotic), $5\mathrm{k}$ (Atari)\\
    & Exploration strategy & Random\\ 
    & Repeat action probability & $0.9$\\ \midrule
    \textsc{Go-Explore} & Exploration strategy & Random\\
    & Repeat action probability & $0.9$\\ \midrule
    \textsc{ICM} & Feature size & $16$\\
    & Networks & $[64, 64]$\\
    & Activation function & ReLU\\ \midrule
    \textsc{Surprise} & Networks & $[64, 64]$\\
    & Activation function & ReLU\\ \midrule
    \textsc{DIAYN} & Discriminator networks & $[256, 256]$\\
    & Activation function & ReLU\\ \midrule
    \textsc{Skew}-Fit & Gradient steps & $100$\\
    & Batch size & $2048$\\
    & Learning rate & $0.01$\\
\end{tabular}
\end{small}
\end{center}
\vskip -0.1in
\end{table*}

The hyperparameters used for the off-policy agent are identical for all algorithms. They are presented in Table \ref{table:hyperparameter_agent}.

\begin{table*}[t]
\caption{Hyperparameters of the off-policy agent. These hyperparameters are identical for all methods and for all experiments. The hyperparameters related to HER relabeling only apply to the methods for which the agent is goal-conditioned (DIAYN, Go-Explore, Skew-Fit and LGE).}
\label{table:hyperparameter_agent}
\vskip 0.15in
\begin{center}
\begin{small}
  \begin{tabular}{llll}
  \toprule
    \textsc{Hyperparameter} & \textsc{SAC} & \textsc{DDPG} & \textsc{QR-DQN}\\ \midrule
    \textsc{Networks} & $[300, 400]$ & $[300, 400]$ & CNN from \cite{mnih2015human}\\
    \textsc{Learning rate} &  $3\times 10^{-4}$ & $10^{-3}$ & $5\times10^{-5}$\\
    \textsc{Learning starts after $N$ timesteps} & $100$ & $100$ & $1\mathrm{M}$\\
    \textsc{Batch size} & $256$ & $100$ & $32$\\
    \textsc{Discount factor ($\gamma$)} & $0.99$ & $0.99$ & $0.99$\\
    \textsc{Polyak update coefficient ($\tau$)} & $0.005$ & $0.005$ & $1.0$\\
    \textsc{Target entropy} & $2.0$ & N/A & N/A\\
    \textsc{Target update every $N$ timesteps} & N/A & N/A & $10\mathrm{k}$\\
    \textsc{$\epsilon$ decreases during $N$ timesteps} & N/A & N/A & $4\mathrm{M}$\\
    \textsc{Initial $\epsilon$} & N/A & N/A & $1.0$\\
    \textsc{Final $\epsilon$} & N/A & N/A & $0.05$\\\midrule
    \textsc{Train every $N$ timesteps} & $1$ & $1$ & $10$\\
    \textsc{Gradient steps} & $1$ & $1$ & $1$\\\midrule
    \textsc{HER sampling probability} & $0.8$ & $0.8$ & $0.8$\\
    \textsc{HER relabeling strategy} & Future & Future & Future\\
    \bottomrule
\end{tabular}
\end{small}
\end{center}
\vskip -0.1in
\end{table*}

For Atari, we mainly use the setting recommended by \citet{machado18arcade}. Like \cite{ecoffet2021first}, we use both sticky actions and start no-ops.

\begin{table*}[t]
\caption{Atari setting.}
\label{table:atari_setting}
\vskip 0.15in
\begin{center}
\begin{small}
  \begin{tabular}{ll}
    \toprule
    \textsc{Parameter} & Value \\ \midrule
    \textsc{Reset on life loss} & Yes\\
    \textsc{Start no-ops} & From $1$ to $30$\\
    \textsc{Action repetitions} & $4$\\
    \textsc{Sticky action probability $\sigma$} & $0.25$\\
    \textsc{Observation preprocessing} & $84 \times 84$, grayscale \\
    \textsc{Action set} & Full (18 actions)\\
    \textsc{Max episode length} & $100\mathrm{k}$\\
    \textsc{Max-pool over last $N$ action repeat frames} &  $2$\\
    \bottomrule
  \end{tabular}
\end{small}
\end{center}
\vskip -0.1in
\end{table*}

\section{On the criticality of cell representation in Go-Explore}
\label{appendix:cell_criticality}

In Go-Explore, similar observations are grouped into cells and each cell encountered is stored in an archive. The cell representation is a critical aspect of Go-Explore. In the \textit{Montezuma's Revenge} environment, a slight variation in cell representation results in an order of magnitude difference in the results.  The cells are used to (1) estimate the density of states encountered in the observation space and sample a target cell against it; (2) divide this goal reaching task into a sequence of subgoals.

We argue that building a cell representation to capture the relevant components of an environment to perform the desired task requires a significant amount of domain knowledge. In general, this cell representation cannot be generalized to other tasks or to other environments. 

\begin{figure*}
\vskip 0.2in
\begin{center}
\begin{subfigure}[b]{0.25\textwidth}
        \centering
        \includegraphics[width=\textwidth]{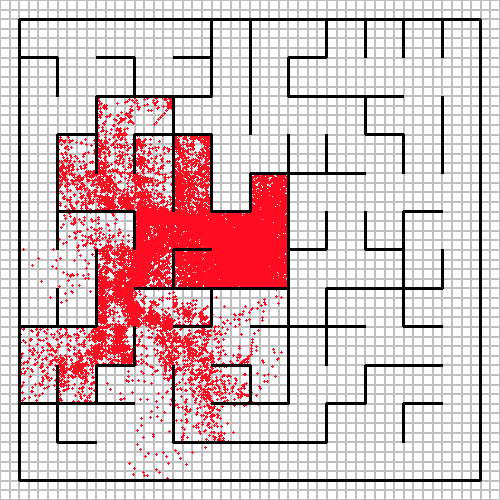}
        \caption{Cell size $= 0.5 \times 0.5$}
    \end{subfigure}
    \begin{subfigure}[b]{0.25\textwidth}
        \centering
        \includegraphics[width=\textwidth]{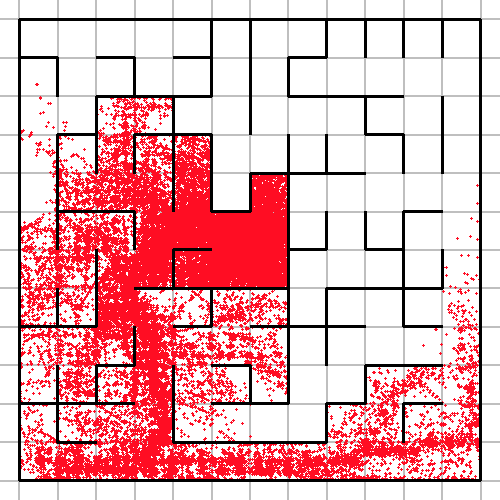}
        \caption{Cell size $= 2.0 \times 2.0$}
    \end{subfigure}
    \begin{subfigure}[b]{0.25\textwidth}
        \centering
        \includegraphics[width=\textwidth]{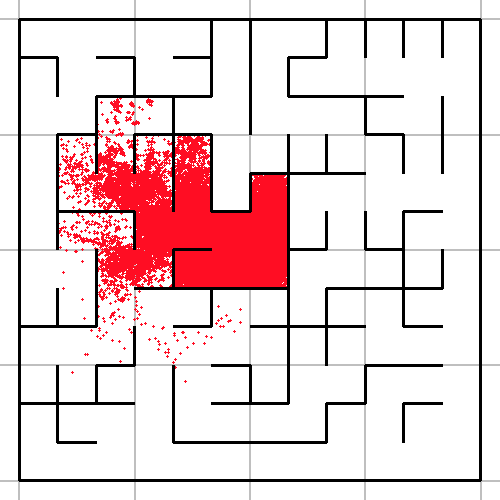}
        \caption{Cell size $= 6.0 \times 6.0$}
    \end{subfigure}
    \hfill
\caption{Go-Explore scene coverage after 100k timesteps. The cell design is represented by the gray grid. We show the results for 3 different cell widths and shift. The red dots represent the visited states.}
\label{fig:cell_criticality}
\end{center}
\vskip -0.2in
\end{figure*}

To support our claim, we present in Figure \ref{fig:cell_criticality} the space coverage in a continuous maze for different cell design. We show that even in this simple environment, a small variation in cell design has a significant impact on the result.

On the left, the cells are small, and the agent must visit each of them. If the agent interacts long enough with the environment, it should eventually explore the whole space. On the right, the cells are large. We can observe some detached areas, because the agent has not visited the cell enough to discover the next one, but enough so that this cell is no longer listed as a target cell.

\section{On the density estimation}
\label{appendix:density_estimation}

Let $s_1, \ldots , s_n$ be a sample of event locations in a $\mathbb{R}^d$-space. Assume that the event location $s$ follows a common distribution with density function $f(s)$. For any sample $s_i$ and $s_j$ in this sample, assume that $D_i(s_j) = ||s_i - s_j||$, denotes the euclidian distance between $s_i$ and $s_j$. For any $k\leqslant n$, let $D_{(k)}(s_i)$ be the distance with $k$-th nearest neighbors of $s_i$ with respect to the euclidian distance.

\cite{kung2012optimal} propose an optimal unbiased estimator $\hat{f}$ for the density:
\begin{equation}
   \hat{f}=\frac{k U^*_k}{k - 1}
\end{equation}
where

\begin{equation}
    U^*_k = \frac{(k-1) }{nC_d D^d_{(k)}}
\end{equation}
and 
\begin{equation}
    C_d = \frac{\pi^{d/2}}{\Gamma (d/2 + 1)}
\end{equation}

Hence we have
\begin{align}
   \hat{f}=\frac{k}{nC_d} D_{(k)}^{-d}
\end{align}

We follow the recommendation of \cite{kung2012optimal}  to take
\begin{equation}
    k = 2n^{1/d}
\end{equation}

\section{Comparing Go-Explore and LGE}
\label{appendix:ge_vs_lge}

The Go-Explore algorithm as presented by \citet{ecoffet2021first} has many components. All these components allow to obtain good results on test environments. In this article, we implement our own version of Go-Explore. We have tried to stick as much as possible to the initial implementation and to improve some aspects.
We keep the essence of Go-Explore, but our implementation is not intended to be equivalent to the initial implementation.
The main goal here is to compare LGE and Go-Explore. Thus, the two implementations differ only in the elements that make them unique. 
To the best of our knowledge, all the composnts that we did not implement are compatible with LGE. It is likely that they improve LGE and Go-Explore in a similar way.
In this section, we describe the implementation of LGE and Go-Explore. We explain their differences if any.

\paragraph{Policy-based Go-Explore}
The initial implementation of Go-Explore distinguishes between the case where the environment can be reset to any desired state and the case where this is not possible.
In this paper, we choose the general setting where the environment can't be reset to any desired state, and we therefore work with the so-called policy-based implementation of Go-Explore.

\paragraph{Exploration after returning}
In the original implementation of Go-Explore, once a cell is returned, exploration proceeds with random actions for a certain number of timesteps. For both LGE and Go-Explore, we set this number of timesteps to 50 for all environments. Note that the agent can interrupt this exploration beforehand if the maximum number of interactions with the environment is reached. \cite{ecoffet2021first} shows that action consistency generally allows for more effective exploration, especially in the robotic environment. For LGE and Go-Explore, we use the same trick: the agent chooses the previous action with a probability of 90\%, and uniformly samples an action with a probability of 10\%.

\paragraph{Cell design}
The original implementation of Go-Explore provides two methods for generating the cell representation.

\begin{enumerate}
\item When the observation is an image, the observation is grayscaled and downscaled. The image produced is the cell representation. The parameters to get this representation (downscaling width and height and number of shades of gray) are optimized during training to maximize an objective function that depends on a target split  factor.
\item When the observation is a vector, each component of the vector is discretized separately by hand before learning.
\end{enumerate}

For all the environments presented in this paper, we use a naive method of cell generation corresponding to a discretization of the observation. The granularity of the discretization is a hyperparameter. The choice of this hyperparameter is crucial, we develop it in more detail in the Appendix \ref{appendix:cell_criticality}.

\paragraph{Goal-conditioning}
In the original implementation of Go-Explore, the agent is conditioned by the cell representation of the goal. We note that this representation can vary during learning, and even in size (see previous paragraph). It is not clear how to structure the agent's network when the size of the input varies during the learning process. 

In our implementation, we choose to condition the agent by the goal observation rather than by the representation of its cell. We also condition the agent by the goal observation in LGE.

\paragraph{RL agent}
In the original implementation of Go-Explore, the goal-conditioned agent is based on the on-policy PPO algorithm \cite{schulman2017proximal}. For both LGE and Go-Explore, we rather chose to use an off-policy algorithm (SAC or DDPG) to use a Hindsight Experience Replay (HER, \citet{andrychowicz2017hindsight}) relabelling, which has shown to perform better in a sparse reward environment. 

\paragraph{Reward}
In the original implementation of Go-Explore, the agent gets $+1$ reward for reaching intermediate cells and $+5$ reward for reaching the final cell of a path. The rest of the time, it receives a $0$ reward.
For both LGE and Go-Explore, following the suggestions of \cite{tang2021hindsight}, we rather choose the following structure for the reward: the agent gets a reward of $0$ for a success (target cell reached for Go-Explore and latent distance with the goal state below the distance threshold for LGE) and $-1$ the rest of the time.
As noted by \cite{eysenbach2019search}, in this setting, an optimal agent tries to terminate the episode as quickly as possible. We therefore set \texttt{done = True} only at the end of the post-exploration, and not when the agent reaches a final state.

\end{document}